# Muscle Vision: Real Time Keypoint Based Pose Classification of Physical Exercises

Alex Moran, Bart Gebka, Joshua Goldshteyn,
Autumn Beyer, Nathan Johnson, and Alexander Neuwirth

Department of Electrical Engineering and Computer Science
Milwaukee School of Engineering
{beyera, gebkab, goldshteynj, johnsonna, morana, neuwirtha}@msoe.edu

## Abstract

Recent advances in machine learning technology have enabled highly portable and performant models for many common tasks, especially in image recognition. One emerging field, 3D human pose recognition extrapolated from video, has now advanced to the point of enabling real-time software applications with robust enough output to support downstream machine learning tasks. In this work we propose a new machine learning pipeline and web interface that performs human pose recognition on a live video feed to detect when common exercises are performed and classify them accordingly. This exercise pose classification pipeline experimentally determines, in a real-time mobile environment, the type of physical exercise being performed from a predefined set of examples. In contrast to existing models that use image recognition directly, our classification model strictly focuses on three dimensional coordinates for key landmarks on the human body, which requires a considerably smaller input size and parameter count for the model than a direct image-based approach. The reduced model size enables a highly performant model that can output live results, even when running in parallel to a landmark extraction model. We derive landmarks by leveraging an existing open-source model built for mobile GPU inference, MediaPipe BlazePose[1], allowing for our model to feed in landmark points and output real-time classifications. The classification model accepts a time series of keypoint positions, allowing our model to consider sequential position dependencies that constitute an exercise. Usage of time series data permits building towards possible future applications in automatic pose validation and correction with regards to the accuracy of individual keypoint movement across the time series. Our web-based interface is highly portable, allowing the model to run on many operating systems and device types, including mobile. The resulting model interface is capable of webcam input with live display of classification results. Our main contributions include a keypoint and time series based lightweight approach for classifying a selected set of fitness exercises and a web-based software application for obtaining and visualizing the results in real time.

# 1. Introduction

Modern day fitness applications often provide interactive exercise guidance, including exercise completion and form correction. To start constructing this type of application, we gathered data by consolidating a small subset of workout videos on four specific exercises. These videos were then run through MediaPipe's BlazePoze[1] which extracted the human figure as 33 key body points. Due to the time series dependency of the data requiring analysis of multiple concurrent frames, we decided to implement a stacked long short-term memory (LSTM) neural network. The time series data allowed for movement of the body to become learned exercises that could then be classified by the network. Our model took in those key body points extracted from running BlazePose on the exercise related subset of the UCF101 dataset [2] categories to be classified. At present our model identifies four physical exercises: pushups, lunges, bodyweight squats and throwing discus. These exercises were selected to represent a diverse range of exercise types by including a broad range of motion and body orientation. Our trained model was uploaded to a custom web application that takes in a live camera feed. The camera feed is piped into BlazePose to get landmarks which our model then uses for reliable classification on the proposed exercise being performed.

Training the LSTM was done in a reasonable timeframe, taking only 6 minutes and 40 seconds to run on MSOE's high performance computing cluster ROSIE using a single NVIDIA T4 GPU, with 8 logical cores allocated from an Intel Xeon Processor with a 2.3 Ghz base clock. Our model's overall accuracy is 95.90% on the four trained classifications of physical exercises. Our experiments suggest that additional training data or augmentation and tests of other models may improve real time classification performance in deployment.

# 2. Related Work

The concept of pose correction using artificial intelligence has been implemented a number of times in the past and requires either manual input of the exercise or an automatic classification of the user's motion into a known exercise. The model we have created performs the latter and follows a similar approach to the work done by Mayorquin et al. [3] which uses a recognition and correction model, YogAI. YogAI uses a two dimensional vector heat map to estimate intended yoga poses as well as give advice on how to better perform a specific pose. Using temporal data YogAI was used to differentiate between standing, squatting, and deadlifting, and was able to count repetitions of the exercise. One primary difference between the model YogAI used was their reliance on OpenPose which, as shown by comparison in the BlazePose[1] paper, is a slower, less real time approach, in terms of inference time.

Work done by Hassan et al. [4] uses manual input to define the exercise that is to be done by the user then assesses the quality of the motion compared to the known correct motion. This takes into account the differences in the user's body shape. This system does not determine the type of exercise performed, but could classify it by comparing the user's motion to all known exercises and choosing the most correct option. The model created by Hassan et al. [4] uses two dimensional input data with a confidence associated with the key points of the body. The feedback given to the user is given after the exercise has been completed, which is in contrast with the real time system we have created.

Chen et al. [5] have created a model that takes a cropped video, creates a generalized two dimensional geometric representation of the key points in the video, then uses one of two models to finally provide feedback to the user. The models used are geometric and dynamic time warping. The model automatically detects the perspective of the user in the video, but this is also not done in real time.

## 3. Model Architecture

### 3.1 Dataset

The pose classification data was retrieved from UCF101 Human Actions dataset [2] which contains 13,320 video clips of categorized activities. From the 101 activities, we selected four physical exercises: pushups, lunges, bodyweight squats, and throwing discus. These activities were selected to test the network on a diverse range of activities. The selected videos are sent through MediaPipe's BlazePoze which extracts the human figure in each frame. The figure is stored as a series of 33 key points representing the human figure. Each point consists of the points visibility and normalized x, y, and z coordinates relative to the detected figure's hips The videos in the UCF101 Human Actions dataset have varying lengths and thus most videos must be padded in order to be fed into the network.

### 3.2 Neural Network Architecture

The model's input tensor shape was determined based on the following factors: the number of training examples $n$, the body landmarks key points $p$, the time steps $t$, and the point location and visibility features $s$, which results in the following mapping into a probability distribution over the exercise classification space:.

$$f(n \times p \times t \times s) \rightarrow \vec{e} \in p(E)$$

In order to feed our model the body landmark key points, the video feed from a laptop or mobile device must first run through the existing model BlazePose, which gives those 33 key landmark points. Each of the points are given as normalized x, y, z, with a confidence value, which are then used as the point attributes for our model. Since there are always four attributes per landmark point and we use a consistent number of landmarks, that dimension of our input tensor is set with a size of 132, as the key points and point attributes are flattened out.

The results of the time series of poses are then evaluated on our model, which gives a confidence for each classification. Eventually, the model output's highest classification confidence is mapped to the exercise being performed. This allows the user to know which exercise is being performed. Under the circumstances that BlazePose is unable to detect the landmark points, we replace nans with the same value used for padding, which is then masked out in the very first layer, which then feeds into the LSTM.

**Figure 1:**

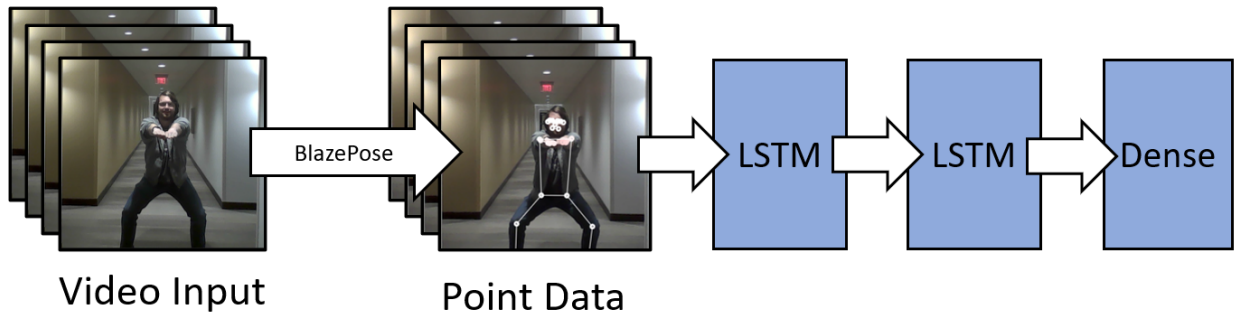

The above figure displays the pipeline used going from a video feed, into the mapped points by BlazePose, then piping those three dimensional points and their confidence values into the stacked LSTM architecture, where the first and second LSTM have a size of 64 nodes each, with the dense layer matching the corresponding output classification size, so with the current four classifications chosen, a size of four. Our model, consisting of the stacked LSTM, has a total of 83,716 parameters, which puts it as a relatively small model in size, where each parameter is a 32 bit floating point number, resulting in a h5 model file size of 692.55kb. The compact size of the model is due to only needing a time series of points, instead of training off of the video feed directly. The small nature of the stacked LSTM best suits our model for mobile deployment.

### 3.2.1 Hyperparameters

Our model used a batch size of 32, and trained for 50 epochs, with a learning rate of 0.0001, using RMSProp for the optimizer, and a categorical cross-entropy loss function. A recurrent dropout of 0.3 was also included for each LSTM layer.

Hyperparameters selection was primarily guided by intuition. The batch size was one of the more carefully chosen values as it needed to be a power of two that was high enough to train in a reasonably quick amount of time, but low enough to support learning for our limited data set size, in order to best take advantage of the hardware.

The number of epochs was adjusted to train up until the point that overfitting may occur, then dialed back to prevent overfitting of the model. The learning rate was set high enough for significant improvements in training accuracy and loss to occur, while not set too high as to fluctuate back below the learned accuracy by a significant amount. RMPProp was chosen as the optimizer after testing out alternate optimizers, including Adam, which showed worse training results, both in terms of time needed to train and the maximum accuracy achieved from training.

The recurrent dropout was meticulously included for each LSTM layer, as the model no longer meets the specifications needed to make effective use of the cuDNN kernel, resulting in a slower training time, by a factor of approximately two, but was included as no adverse effects were seen in the trained model, which had a higher accuracy compared to training without recurrent dropout.

## 4. Experiments

Model quality was evaluated primarily on the accuracy and loss observed while training, with the validation accuracy as the most important aspect of the results, in addition to deployment testing. Since the evaluation of the model is more subjective when looking at the results of deployment, in addition to the multitude of potential deployment options for this model, our experiments focused on the overall accuracy, and the validation set used. Our model achieved an overall accuracy of 95.90% on our four classifications. Table 1 below shows the breakdown of accuracies by classification.

**Table 1:**

|  | Body Weight Squats | Lunges | Push Ups | Throwing Discus |
|---|---|---|---|---|
| Total Accuracy [%] | 96.43 | 94.49 | 99.00 | 94.35 |
| Validation Accuracy [%] | 96.67 | 95.83 | 92.86 | 95.83 |

The above table holds the accuracy for the entire data set as well as the validation set. The rationale for including the results of the trained upon data is to highlight the lack of overfitting, as the majority of the accuracies were greater on the validation set than on the trained upon set.

### 4.1.1 Confusion Matrix:

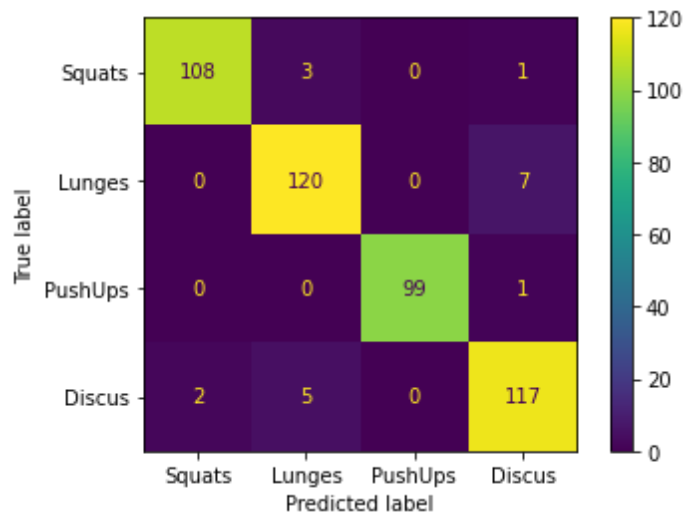

The above figure shows a confusion matrix for the predictions made on the entire data set used. Although the total counts are not equal across the categories, resulting in a skew against

PushUps when looking at the color gamut key, the confusion matrix is still beneficial as it provides the numerical count of true positive predictions, in addition to what negative predictions were. Table 1 highlights the PushUps as the most correctly classified category, where the above confusion matrix does not convey the same message until further inspection, where we can see that only one push up video was misclassified as a discus throw.

**4.1.2 Confusion Matrix of Test Set:**

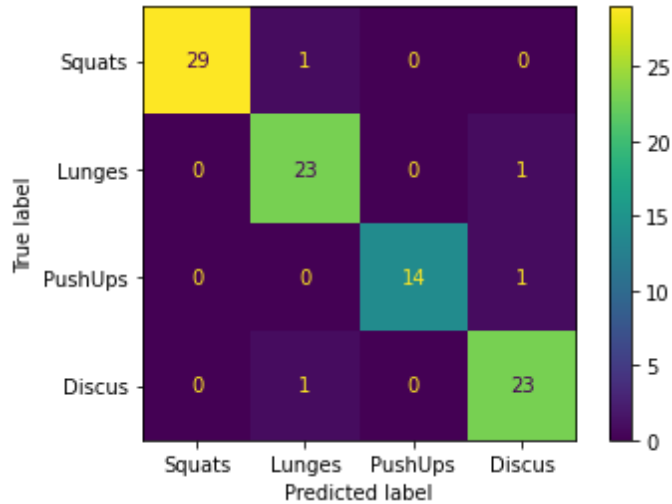

The above confusion matrix gives further insight to where our model is potentially suffering the most, however again we see the imbalance between classifications as the data set is not split to evenly allocate input across the classifications, in part due to the limited data set. In practice push ups were classified most reliably compared to the other physical exercises, and yet because of one misclassification here the validation data appears skewed against the performance of classifying pushups.

# 5. Applications

The goal of this model was to create something that would effectively be able to classify pushups, body-weight squats, and lunges. This classification was then combined with a web-based front-end that would allow for a user to classify the exercise that they are performing via the use of a webcam or the built in camera on a smartphone. This can also be used with remote cameras and allows for classification of the exercise further away from where the calculation is being performed.

As a result of this model, the exercise that is being performed can be ascertained with a high (90%+) accuracy according to the validation model. This would allow for a personal trainer or person who goes to the gym to accurately track which exercise is performed, as well as when it is performed. This would allow for someone to also track how long each exercise repetition is being performed in addition to the total workout duration.

The logging ability of the model and its corresponding web application will allow for someone to reflect on the workout that has been performed (specifically in a circuit-training type plan) and adjust their future workouts based on their categorized performance.

While a high-end GPU and server-grade hardware were used in the training of the model, the running or evaluation of the model is significantly faster and can be run in real-time on almost any modern consumer device. As a result, this allows for a cross-platform deployment that will still have an acceptable level of performance for the user of the model.

### 5.1 Deployment

The trained model was exported from keras as an h5 file, then transformed into a TensorFlowJS model which the frontend uses in tandem with MediaPose. In order to account for video length bias, only the most recent 8 frames are imputed into the LSTM. The remaining data is masked to preserve data size. The current deployment supports real time inference on laptops, with further testing needed to support alternate mobile devices.

Ease of use is a factor to consider, as in our tested method of deployment, through the use of a web interface, the video feed remained reliable with minimal performance impact caused by inferences made. Running on laptops with an Intel i5-10200U, with a boost clock speed reliably above 2.3-3.0 Ghz we were able to realize a usable framerate, which varies depending on the quality mode of the BlazePose model. For the use of generating training data and deployment our model takes in the inputs from the medium BlazePose quality setting, resulting in a framerate between the lite and full models, hovering around the low to mid 20's for frames per second. The frame rate is high enough to realize the benefits of real time inference, with the resultant classification displayed in plain text to the user.

### 5.2 Future Work

Given the high validation accuracy of the model, and the real time performance on laptops, further development would focus on implementing full mobile phone and tablet support. The model given its current state is built to run on mobile devices as it is a rather compact model, and would only need modifications made to a front end to ensure mobile support. Our model also lays the foundations for pose correction, as a lightweight classification model may be incorporated into a future pipeline for correcting exercise form in real time.

## 6. Conclusion

While misclassifications occur, it is important to inspect the nature of misclassifications, as mentioned when discussing the accuracy of the validation set and on the dataset subsection as a whole. The confusion matrix shown above under Section 4.1.1 gives insight into the most commonly performed misclassification inferences. The high values along the diagonal, which represent true positives are shown relatively strong, but looking at the negative values we can see that the two classifications most commonly mixed up by our model are the luges and discus throws. In our case seven lunges were predicted as discus throws and five discus throws predicted as lunges. While the two exercises are not obviously similar in nature, the way that lunges and squats may be, from the similarity of leg movement, there are a number of factors that could be throwing off the classification. These factors include potential low visibility of body landmarks, a lack of detection of the face, i.e. when a lunge is performed perpendicular to the camera, and while a thrower is spinning, et cetera.

Real time performance of classifications performed by our lightweight but robust model means deployment is already possible and has been proven to run inferences with minimal effect on the frame rate of the video feed. Our model displays potential as an intermediary for subsequent models further along the pipeline.

# References

**[1]** Bazarevsky V., Grishchenko I., Raveendran K., Zhu T., Zhang F., & Grundmann M. (2020). BlazePose: On-device Real-time Body Pose tracking. arXiv:2006.10204 [Cs]. https://arxiv.org/abs/2006.10204

**[2]** Khurram Soomro, Amir Roshan Zamir and Mubarak Shah, UCF101: A Dataset of 101 Human Action Classes From Videos in The Wild., CRCV-TR-12-01, November, 2012.

**[3]** Salma and Terry, "Yogai: Smart personal trainer," Feb 2019. [Online]. Available: https://www.hackster.io/yogai/yogai-smart-personal-trainer-f53744

**[4]** Hassa H., Abdallah B., Adallah A., Abdel-Aal R., Numan R., Dawish A., El-Bhaidy W. (2020). Automatic Feedback For Physiotherapy Exercises Based On PoseNet.

**[5]** Chen S., Yang R. (2020) Pose trainer: Correcting exercise posture using pose estimation. arXiv:2006.11718v1 [cs.CV]
https://arxiv.org/abs/2006.11718